\documentclass[letterpaper, 10 pt, conference]{ieeeconf}  
\IEEEoverridecommandlockouts 
\usepackage{subfigure}
\usepackage{color}
\usepackage{graphicx}    
\usepackage{float}
\usepackage{multirow}
\usepackage{algorithm}
\usepackage{algpseudocode}
\usepackage{amsmath}
\usepackage{algorithmicx}
\usepackage{algpseudocode}
\usepackage{hyperref}
\usepackage{threeparttable}
\usepackage{booktabs}

\usepackage{indentfirst}
\usepackage{amsmath}
\usepackage{cite}
\usepackage{pdfrender}
\usepackage{amssymb}
\usepackage{amsbsy}
\usepackage{bbding}

\usepackage{color}

\title{\LARGE \bf
Learning 6-DoF Task-oriented Grasp Detection via 

Implicit Estimation and Visual Affordance
}

\author{Wenkai Chen$^{1*}$,  Hongzhuo Liang$^{1}$, Zhaopeng Chen$^{2}$, Fuchun Sun$^{3}$ and Jianwei Zhang$^{1}$
\thanks{$^{1}$Technical Aspects of Multimodal Systems (TAMS), Department of Informatics, Universit\"{a}t Hamburg}
\thanks{$^{2}$Agile Robots AG}%
\thanks{$^{3}$Beijing National Research Center for Information Science and Technology (BNRist), State Key Lab on Intelligent Technology and Systems, Department of Computer Science and Technology, Tsinghua University}
\thanks{*Corresponding author to provide e-mail: wchen@informatik.uni-hamburg.de}%
}

\begin{document}
\maketitle
\thispagestyle{empty}
\pagestyle{empty}

\begin{abstract}
Currently, task-oriented grasp detection approaches are mostly based on pixel-level affordance detection and semantic segmentation. These pixel-level approaches heavily rely on the accuracy of a 2D affordance mask, and the generated grasp candidates are restricted to a small workspace. To mitigate these limitations, we firstly construct a novel affordance-based grasp dataset and propose a 6-DoF task-oriented grasp detection framework, which takes the observed object point cloud as input and predicts diverse 6-DoF grasp poses for different tasks. Specifically, our implicit estimation network and visual affordance network in this framework could directly predict coarse grasp candidates, and corresponding 3D affordance heatmap for each potential task, respectively. Furthermore, the grasping scores from coarse grasps are combined with heatmap values to generate more accurate and finer candidates. Our proposed framework shows significant improvements compared to baselines for existing and novel objects on our simulation dataset. Although our framework is trained based on the simulated objects and environment, the final generated grasp candidates can be accurately and stably executed in the real robot experiments when the object is randomly placed on a support surface.
\end{abstract}

\section{Introduction}
Recently, task-oriented robotic grasping and manipulation have received more and more attention from the robotics community\cite{haschke2005task,li1988task,fang2020learning}, which aims to generate different robotic actions and interactions for the same object representing of a potential scenario. The outcome of task-oriented robotic motion can benefit a robot's ability to understand the semantic context of objects better. For example, traditional robotic grasping detection methods (pixel-based and point cloud-based) all generate random grasp candidates around the target object. However, these methods lack the understanding of object context (such as global and local texture).

To cope with this limitation, some researchers introduce the concept of affordance into the robotic field, which plays a key role as mediator, organising the diversity of possible perceptions into tractable presentations that can support reasoning processes to improve the generalization of tasks \cite{mar2015self,stoytchev2005behavior}. However, most of the current mainstream affordance-based robot grasping methods require prior pixel-level target detection and semantic segmentation \cite{chu2019learning,nguyen2017object,sawatzky2017adaptive,xu2021affordance,chu2019toward}, where grasping candidates are generated after obtaining the target object part. Nonetheless, this kind of grasp detection strategy cannot essentially couple with the contextual information of objects, and the pose of pixel-based grasp is at a limited dimension. To combine 6-DoF grasp detection with affordance knowledge, some works are also proposed to use the observed point cloud as the model input recently\cite{jiang2021synergies,qian2020grasp,murali2020same,liu2020cage}. These methods either rely solely on a generative model to generate target region grasping, or use mask to assist grasping detection, which cannot achieve a great trade-off between grasp quality and generalization ability.

\begin{figure}[!t]
\centering
\includegraphics[width=0.48\textwidth]{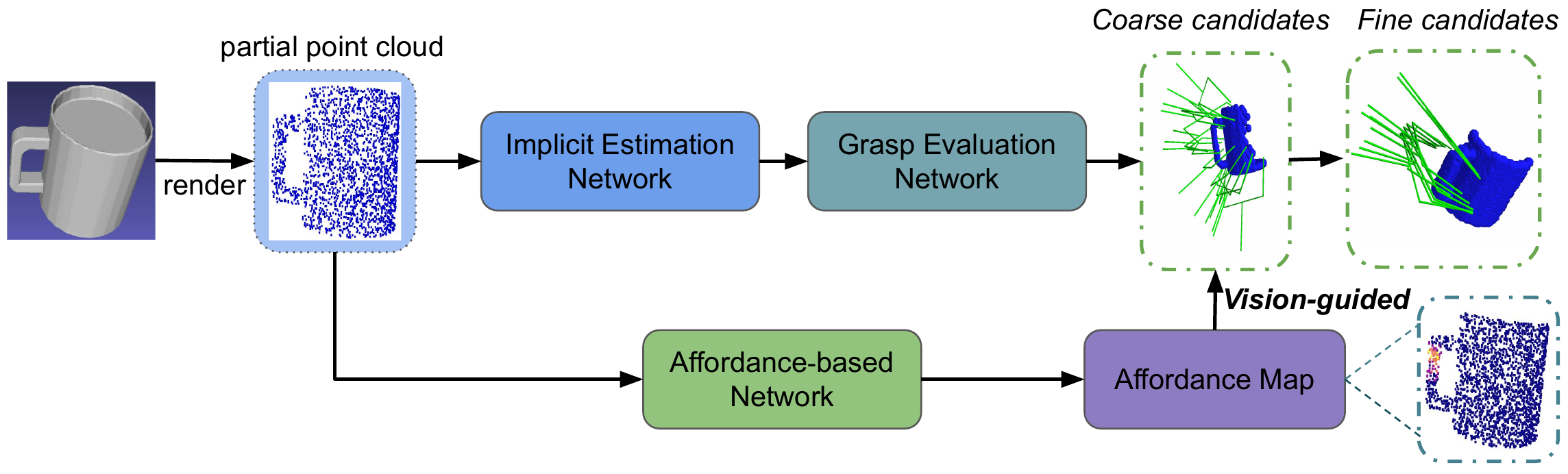}
\caption{Overview of our 6-DoF task-oriented grasp detection framework for affordance-based robotic grasping. The observed point cloud from the RGB-D camera are sampled as input $\mathcal{P} \in \mathcal{R}^{N\times3}$. It will pass into two parallel modules, the grasping affordance module and the visual affordance detection module. The first module finally outputs coarse grasp candidates $\mathcal{G_C}$ in the form of $SE(3)$ and corresponding confidence scores; the other module outputs a 3D heatmap $\mathcal{H}^N$  where the values of each affordance label are predicted. Finally, the visual affordance map values combine with confidence scores to guide the coarse grasp candidates $\mathcal{G_C}$ in becoming more accurate and fine. And fine grasp candidates $\mathcal{G_F}$ will be executed in the real robotic experiments.}
\label{Overview}
\vspace{-0.5em}
\end{figure}

As illustrated in Figure~\ref{Overview}, we present a novel 6-DoF task-oriented grasp detection framework for affordance-based robotic grasping task. Specifically, the input of our framework is the partial object point cloud captured by the RGB-D camera. Our framework consists of two modules: grasping affordance detection module and visual affordance prediction module. Motivated by the grasp generation approach based on the generative model (VAEs)\cite{mousavian20196}, it achieves great grasp generation results for some clearly outlined objects like a mug, scissor and bottle. Due to the complex data distribution and geometry structures of points, we postulate that the uni-modal distribution assumption can be violated in multi-affordance grasping generation tasks. Especially the generated grasp poses from point-wise features can vary vastly within different affordances. Thus, we design each task with an implicit representation to better represent the complex distribution. Otherwise, a grasp evaluation network is designed to evaluate the generated coarse grasp candidates. Both networks are trained in our self-constructed affordance grasp dataset. For the visual affordance prediction module, inspired by the work of 3D affordanceNet benchmark~\cite{deng20213d}, we designed an attention-aware bilinear feature learning network to capture the geometric dependencies and semantic correlations by learning point features and edge features, simultaneously. Sequentially, the predicted affordance map guides coarse grasp candidates to centrally distribute around the spatial region of the maximum map value, which could significantly improves the accuracy and stability of the generated grasps.

The main contributions of this paper can be summarized as:
\begin{itemize}
\item Based on the work of ACRONYM\cite{eppner2021acronym}, we introduce an affordance grasp dataset where each successful grasp of each object is annotated as a task-oriented label. Each grasp is evaluated by the simulated engine. 
\item An implicit multi-stream network is proposed to generate diverse affordance-based grasp candidates directly, showing a better performance than the VAEs model. 
\item To use the spatial context of objects, we design an attention-aware bilinear feature learning network and first introduce 3D visual affordance to real robotic grasping, which effectively guides coarse grasp candidates to become more affordance-centric and finer. We also demonstrate that these grasps generated from simulated objects can be transferred to real world.
\end{itemize}
\section{RELATED WORK}
\textbf{Deep Visual Grasping Detection}
Many deep leaning-based approaches have been proposed to tackle the grasp detection problem using full supervision, \emph{i.e.}, pixel-wise or point-wise ground truths.~\cite{lenz2015deep} proposed a two-step cascaded system to combine image detection and grasp generation by training different deep neural networks.~\cite{redmon2015real} adopted a single-stage regression model to graspable bounding boxes to improve grasp detection performance. However, these earlier grasp detection methods are mostly based on pixel images, causing generated grasp poses to be at a limited state and action space like top-to-down grasps. Recently, 6-DoF grasp detection methods have become the mainstream for robotic grasping tasks. Based on point cloud,~\cite{ten2017grasp} proposed a grasp pose detection (GPD) algorithm to generate grasps using a sampling strategy. Based on GPD alorithm,~\cite{liang2019pointnetgpd} used PointNet\cite{qi2017pointnet} to process point cloud for grasping evaluation, achieving to produce diverse grasp candidates of high quality. However, we find that many generated grasps from GPD lack understanding of the surface and contours of objects on a physical interaction level, and it will cause lots of failures on larger and clearly outlined objects\cite{Chen2022}. To address this problem, \cite{mousavian20196} introduced variational autoencoder (VAEs) to generate grasp samples from point cloud, where generated successful grasps embody a certain understanding of the target object. To further improve the physical interaction between object point cloud and the generated grasps, we employ an implicit estimation model to end-to-end generate grasps of objects for different semantic parts without extra visual preprocessing like segmentation and transformation. 

\textbf{Affordance-based Robotic Grasping}
~\cite{gibson1979theory} firstly introduced the concept of affordance, which characterizes the functional properties of objects. Widely accepted by the robotics community, the goal of affordance learning is to reason different physical and contextual meaning of objects~\cite{zhu2014reasoning,do2018affordancenet}. It's very challenging to achieve affordance learning on robotic scene, researchers proposed many approaches to learn different object parts through pixel-wise or point-wise feature and corresponding semantic information~\cite{nguyen2017object,myers2015affordance,song2015task,fang2020learning}. 

For the robotic grasping based on pixel-wise affordance,~\cite{vahrenkamp2016part} labeled different object parts as semantic information to guide the robot to grasp though it can only be applied to similar objects. To obtain a better generalization ability,\cite{rezapour2019towards} proposed to use data-driven approach to accomplish part-based affordance detection, which demonstrates robot could execute successfully after detecting the pixel-wise affordance. Furthermore,~\cite{liu2020cage} proposed a context-aware grasping engine (database), which combines part affordance, part material, and tasks to train a semantic grasp network. That improves the relationship between grasping and objects though it cannot generate diverse grasp candidates automatically.
On the basis of traditional pixel-wise part segmentation, \cite{xu2021affordance} introduced an extra keypoint detection module, whose predictions consists of position, direction, and extent, guiding a more stable grasp pose. However, the problem of these pixel-based part affordance methods is that 6-DoF grasp detection is hard to embed in it, causing generated robotic grasp candidates in a very restricted workspace.

Only recently, some works started to study affordance-based learning on observed point clouds by extending semantic segmentation methods to the point-wise level.~\cite{hjelm2015learning} used a demonstration method to learn tasks, specifically grasping based on visual point cloud. ~\cite{ardon2019learning} proposed a grasp affordance patch mapping method to generate optimal grasping region and then execute grasp while whole execution process is cumbersome.~\cite{jiang2021synergies}
proposed a GIGA framework to use implicit representation,  jointly learning grasp affordance and 3D reconstruction. It achieves a great state-of-the-art grasping performance, while the weakness is that each object is related to a single affordance. Furthermore, ~\cite{murali2020same} collected a TaskGrasp dataset by scanning real object point cloud and divided each object into diverse tasks, which also introduced graph knowledge to help task-oriented grasping generation. However, the grasps in this dataset are annotated purely through the geometric shape of the object, and assumes that each affordance is true manually without considering the true context of object. In this paper, we also construct an affordance dataset where each grasp is evaluated through the simulation engine. Moreover, we propose to combine task-oriented 6-DoF grasp detection with 3D visual affordance of the object together to achieve a better grasp performance and then transfer the simulated grasps to real objects.

\section{PROBLEM FORMULATION}
We consider a setup consisting of a robotic arm with parallel-jaw grippers, an RGB-D camera and an object on a planar tabletop to be grasped. A single-view depth map is captured by the RGB-D camera to convert into a 2.5D partial point cloud $\mathcal{P} \in \mathcal{R}^{N\times3}$ and then passes into the pipeline. For simplicity, all spatial quantities are in camera coordinate frames.

Our pipeline consists of two models: the Grasping Affordance Detection Model and the Visual Affordance Prediction Model. The first model aims to learn a posterior distribution $\mathcal{D((G(T)^*)|P)}$, where $\mathcal{P}$ is the input partial point cloud and $\mathcal{G(T)^*}$ represents successful grasps of different mini-task actions $\mathcal{T} \in (0,M) $, such as wrap, grasp, pour, and cut, where $M$ is the total number of mini-task categories.  This model outputs coarse 6-DoF grasp detection candidates $\mathcal{G_C}$ and associated confidence scores $\mathcal{S_O}$. Furthermore, the function of the second model is to predict a 3D visual affordance map $\mathcal{S_M} \in [0, 1]^{N}$ for different mini-task actions $\mathcal{T}$, which are combined with original grasp confidence scores $\mathcal{S_O}$ to obtain fine grasp candidates $\mathcal{G_F}$. Each generated grasp $G_i\in\mathcal{ (G_C, G_F)}$ is denoted as $(R, T)\in  SE(3)$. We trained our grasp framework by randomly rotating the objects with different affordance in a simulated rendering environment, where final generated grasps $\mathcal{G_F}$ are defined according to the object reference frame whose axes are parallel to the camera. Finally, the $\mathcal{G_F}$, representing fine successful grasps of a certain task affordance, will be transformed into the camera coordinate frame to be executed in the real robot experiments.

\section{Multi-task Grasp Generation Framework with Implicit Estimation and Visual Affordance}
In this section, we firstly collect an affordance grasp dataset based on the simulation and implementation of the above-mentioned two modules, and then describe how they are jointly trained to learn successful grasps from partial object point cloud with different task-oriented affordances. 

\begin{table}[b!]
\caption{Statistic of object number and corresponding affordance categories in the affordance grasp dataset}
\resizebox{\linewidth}{!}{%
\begin{tabular}{c|c|c|c|c|c|c}
\hline
Object     & Mug                                                                  & Bottle                                                         & Knife           & Hat         & Bowl        & Scissor    \\ \hline
Affordance & \begin{tabular}[c]{@{}c@{}}grasp, wrap,\\ pour, contain\end{tabular} & \begin{tabular}[c]{@{}c@{}}grasp, wrap,\\ contain\end{tabular} & grasp, cut/stab & grasp, wear & grasp, wrap & grasp, cut \\ \hline
Number     & 60                                                                   & 33                                                             & 42              & 8           & 52          & 8          \\ \hline
\end{tabular}%
}

\label{tab:dataset}
\end{table}

\subsection{Affordance Grasp Dataset Construction}
Inspired by the 3D AffordanceNet~\cite{deng20213d} and ACRONYM dataset~\cite{eppner2021acronym}, we focus on constructing a new dataset for task-oriented grasping based on simulated ShapeNet~\cite{chang2015shapenet} objects. We choose the ACRONYM dataset as our grasp prototype because it is a large and well-established dataset for robot grasp planning, which in total contains more than 17M parallel grasps and diverse objects from different categories. Moreover, each grasp in this dataset is evaluated and then judged as a successful or failed one through a physics simulator. As shown in Table~\ref{tab:dataset}, we exclude many object categories because they lack affordance meaning and could not be applied to household robotic scenarios.

\begin{figure}[!t]
\centering
\includegraphics[width=\linewidth]{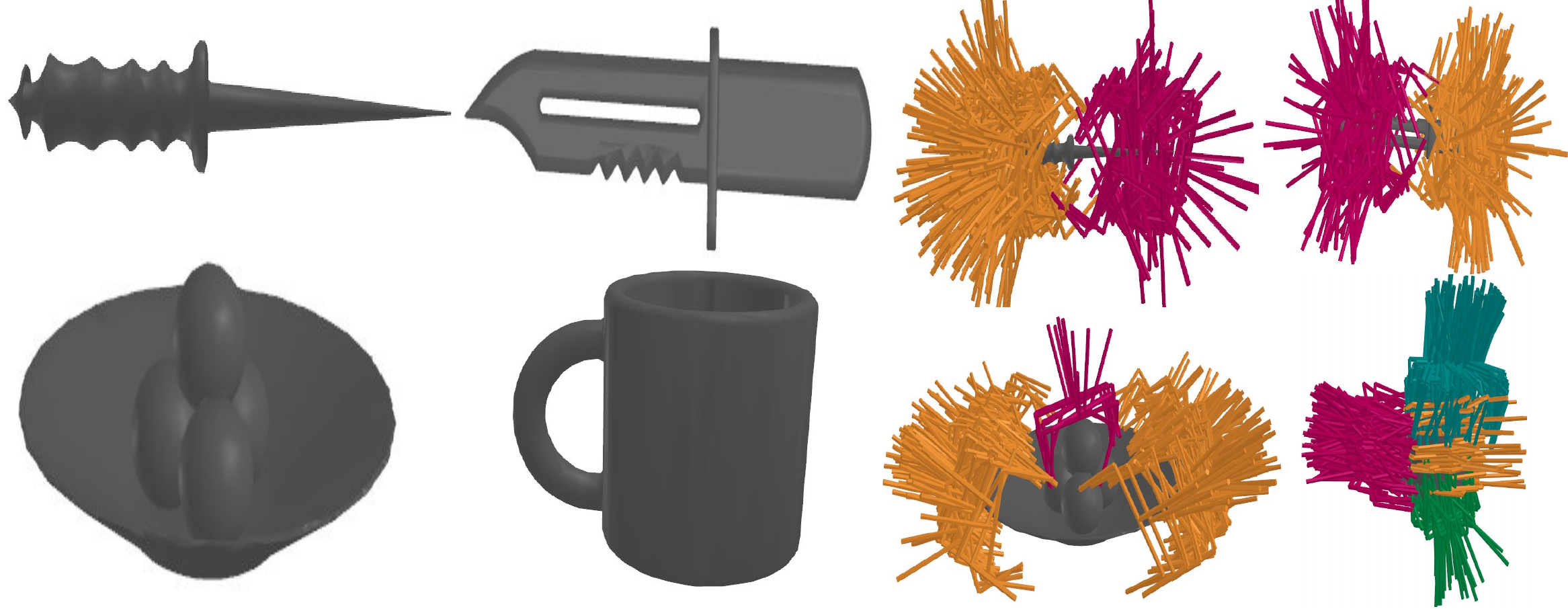}
\caption{Visualization of our partial affordance grasp dataset. (Left) 3D model of ShapeNet objects. (Right) All markers represent successful grasps and different colors of markers indicate different grasp tasks.
}
\label{dataset}
\vspace{-0.5em}
\end{figure}

Rather than using different object parts as different affordance representations ~\cite{deng20213d,liu2020cage,xu2021affordance} , we annotate all successful grasps of selected objects with different affordance labels. For example, \emph{grasp} affordance in the mug instance means all successful grasps around the mug handle while \emph{pour} affordance means all successful grasps around the upper mug rim. During our annotating process, the affordance label number for different objects of same categories are assumed to be the same though we find the distribution of successful grasps of a few objects is not similar. Taking the mug as example, all successful grasps of some mugs with special shape can only be divided into two kinds of affordance types. As a result, we use a constant grasp pose value to indicate successful grasp for the other two kinds of affordance types. All failed grasp candidates of selected objects existing in~\cite{eppner2021acronym} are also reserved in our dataset as negative grasp samples. Otherwise, we transform all objects and corresponding grasps into a uniform coordinate frame, which is conducive to the training of two subsequent modules. Figure~\ref{dataset} visualizes different affordance results of selected objects in our dataset. Finally, our affordance grasps dataset consists of 203 household objects from 6 categories, and more than 100K successful grasps are selected and annotated into 6 common tasks in our daily life.

\subsection{Grasping Affordance Detection Module}
In Figure~\ref{Overview}, our grasping affordance detection module consists of two sub-networks: an implicit estimation network and a grasping evaluation network. Firstly, based on the proposed dataset, the object point cloud is captured by the camera by rotating the object at a random pose in the rendering environment, and each point cloud is sampled to 2048 points through farthest point sampling (FPS). Then the implicit estimation network takes the partial point cloud as input and outputs diverse grasp candidates corresponding to its affordance label. On the other side, the grasping evaluation network takes different grasps as input and learns a classifier to recognize success and failure.
Finally, coarse grasp candidates of each affordance label will be obtained when the generated grasps from the implicit estimation network are input into the trained grasping evaluation network.
Below, we present details of these two sub-networks.

\begin{figure}[t!]
\centering
\includegraphics[width=\linewidth]{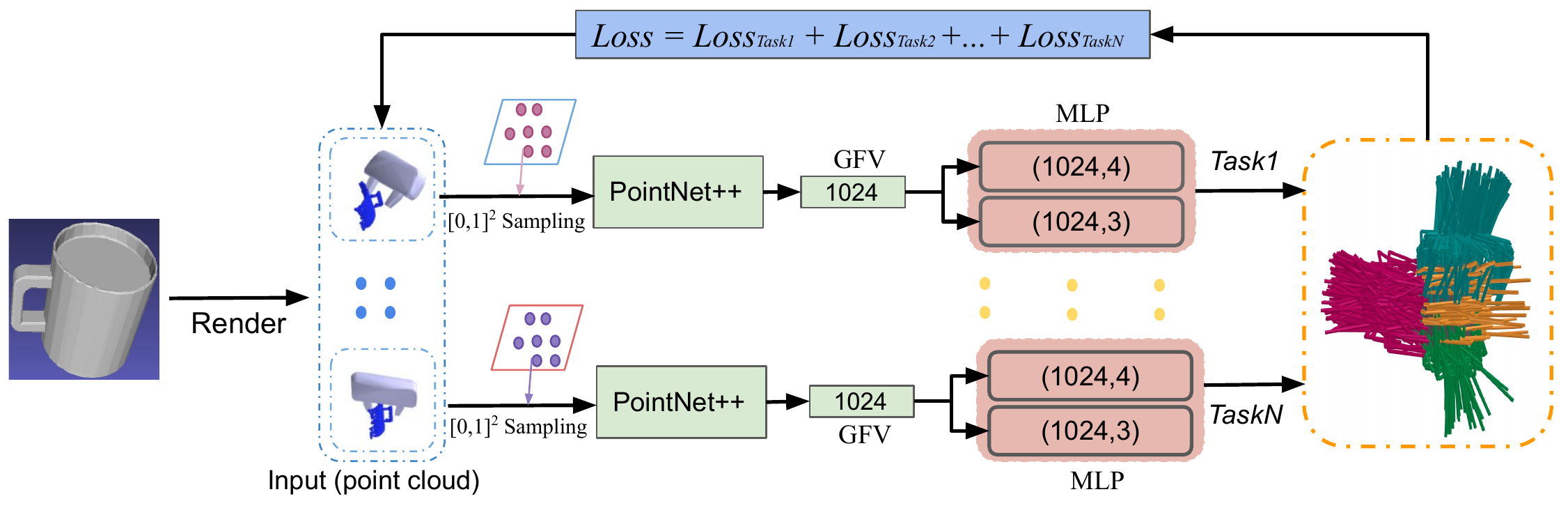}
\caption{The architecture of our proposed implicit multi-stream estimation model.
}
\label{implicit}
\vspace{-0.5em}
\end{figure}

\textbf{Implicit Estimation Network:}
Generative modelling is a cornerstone for machine learning, which has been widely used in the 2D vision field, like image tampering and image compositing. In previous 3D point cloud-based robotic research~\cite{mousavian20196,murali2020same}, variational autoencoders (VAEs) is commonly chosen to 
numerous grasp candidates. However, an accurate VAEs model usually needs a prior partition function to predict the distribution of ground truth, like mixture of Gaussian, hidden Markov or Boltzmann machine. Especially when we need to predict $SE(3)$ grasps of different affordance tasks simultaneously, it is challenging to sample from these models. On the other hand, using an implicit model is a more natural way in terms of the sampling strategy~\cite{grover2018flow,li2018implicit,goodfellow2014generative}, which can be simply expressed by the following sampling procedure:

1. Sample $\zeta \sim \mathcal{M}(0,I)$

2. Return $\chi:= \mathcal{N}(\zeta)$

\noindent Where $\mathcal{M}$ is a latent distribution and $\mathcal{N}$ is a highly expressive function approximator, usually replaced by a neural network.

To encode both affordance and geometry from partial point cloud, we use implicit maximum likelihood estimation method \cite{li2018implicit} to predict grasp poses. As shown in Figure~\ref{implicit}, we propose a multi-stream neural network architecture for jointly predicting $SE(3)$ grasps of different affordance tasks. Each stream represents different affordance label of certain object. At each network stream, the sampled point cloud $\mathcal{P}^{2048\times3}$ is concatenated with a latent indicator $\mathcal{L}$ and then is input into the PointNet++~\cite{qi2017pointnet++} architecture to extract spatial information between the point cloud and the potential grasp pose. After that, a 1024-d generalized feature vector (GFV) can be obtained. We parametrize each GFV with separated small full-connected layers. The output rotation and translation values of the target grasp pose are expressed as:
\begin{equation}
    R^\mathcal{T}_\mathcal{P} \longrightarrow {[quat_1,quat_2,quat_3,quat_4]}
\end{equation}
\begin{equation}
    T^\mathcal{T}_\mathcal{P} \longrightarrow {[X,Y,Z]}
\end{equation}

\noindent where $\mathcal{T}$ and $\mathcal{P}$ separately represent the affordance label and the object point cloud, and rotation values are predicted as a form of quaternion.

\textbf{Grasping Evaluation Network:}
Similar to the evaluation method of~\cite{mousavian20196}, we choose to combine all grasp poses $(\mathcal{G}_s,\mathcal{G}_f)$ with object point cloud $\mathcal{P}$ as the input of the network, where a gripper point cloud corresponding to each grasp pose is used to approximate the real gripper. An extra binary value is also used to judge whether the point from the combined point cloud belongs to the object or gripper. Furthermore, like the implicit estimation network, we still use PointNet++~\cite{qi2017pointnet++} to explore the spacial relationship between object point cloud and gripper point cloud. The output module consists of three full-connected layers [1024, 512, 256] and a final sigmoid layer. Finally, according to the binary ground truth label (success or failure), it is easy to train a classifier to predict the successful probability of each input grasp. After finishing the evaluator training, this model is used to deal with output results from implicit estimation network, which can guarantee final grasp candidates are all successful.

\subsection{Visual Affordance Prediction Module}
\begin{figure}[t!]
\centering
\includegraphics[width=\linewidth]{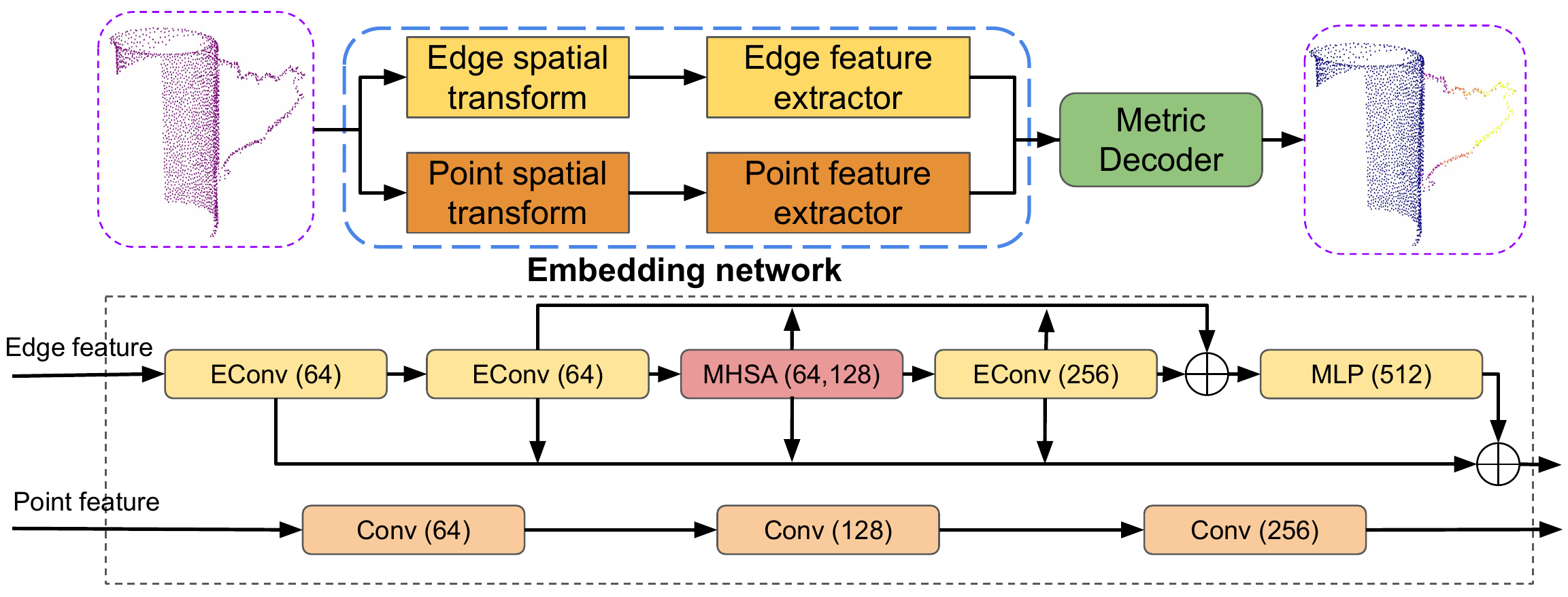}
\caption{The architecture of our proposed visual affordance network, where EConv is the EdgeConv layer and MHSA is the multi-head self-attention module.
}
\label{visual}
\vspace{-0.5em}
\end{figure}

Figure~\ref{visual} illustrates our attention-aware visual affordance network architecture, which consists of two main components: embedding network and metric decoder. The embedding network is the most important part in our network since the performance of the metric decoder relies on learned embedding space. We expect this embedding work to realize two critical functions: 1) to encode the geometric relationship of the local region, especially for different affordance parts. 2) to encode global semantic information based on global context. 

Based on this idea, we design an attention-aware bilinear learning framework that incorporates the point features and edge features to extract local and global semantic information. In particular, we adopt the PointNet~\cite{qi2017pointnet} and DGCNN~\cite{wang2019dynamic} as our backbone to extract different semantic features respectively. Based on PointNet, three sequential convolutional layers (Conv(64, 128, 256)) are used to produce global semantic features.~\cite{wang2019dynamic} proposed a dynamic graph network architecture, which could effectively output local geometry features from the first EdgeConv layer (EConv(64)) and global space features from the final multi-layer perceptrons layer (MLP(512)). To further obtain local correlations for each affordance part, the multi-head self-attention (MHSA(64, 128)) module is applied to generate more semantic features, encouraging point-wise features to aggregate with global context for affordance-meaning point cloud. After obtaining the point feature and edge feature, they are concatenated together to input into the metric decoder, which consists of a stack of multi-layer perceptron layers. It finally predicts a probability distribution based on the partial point cloud corresponding to different affordance tasks:
\begin{equation}
    \mathcal{H}^N = \left\{ H^N_1,...,H^N_M | H^N_i = map_i(\mathcal{P}) \right\}
\end{equation}
\begin{equation}
    H^N_i = \left\{ h^0_i,...,h^{N-1}_i \right\}, h^j_i \in [0, 1]
\end{equation}

where $H^N_i$ is the predicted point cloud map values of affordance label $i$. In our network model, the parameters of  $N$ and $M$ represent the sampled point cloud number and affordance task number, respectively.

\subsection{Training}
The grasp affordance module is trained based on our proposed grasp affordance dataset, and visual affordance prediction module is trained on the 3D AffordanceNet dataset~\cite{deng20213d}, which causes two kinds of training loss. For the grasp affordance loss, we firstly compute minimum L1 loss between any predicted grasp pose $\mathcal{G_C} = [quat_1,quat_2,quat_3,quat_4, X, Y, Z]$ with the ground-truth grasp poses $\mathcal{G(T)^*}$ for each affordance task $\mathcal{T}$. To simplify the computational complexity, each grasp pose is transformed to 7 control points representing a gripper to make training simpler. Thus, the loss function of the implicit estimation network is denoted as $\mathcal{L}_i$:
\begin{equation}
    \mathcal{L}_i = \frac{1}{M} \sum_{i}^{M}\left(min(\frac{1}{k}||\mathcal{G_C}-\mathcal{G(T)^*||}_1)\right)
\end{equation}

Moreover, for the loss function of the grasping evaluation network, we adopt the standard binary cross-entropy loss between the predicted grasp status $\overset{*}{o} \in [0, 1]$ and the ground-truth grasp label $o \in \left\{0, 1\right\}$ (0 for failure, 1 for success). Thus, the loss of the grasp evaluator is denoted as $\mathcal{L}_e$:
\begin{equation}
    \mathcal{L}_e = -(olog(\overset{*}{o})+(1-o)log(1-\overset{*}{o})
\end{equation}

As visual affordance loss, the same training target in benchmark~\cite{deng20213d} is adopted:
\begin{equation}
    \mathcal{L}_v(\overset{*}{m},m) = W_1*L_{ce}(\overset{*}{m},m)  + W_2*L_{dice}(\overset{*}{m},m) 
\end{equation}

\noindent where $\overset{*}{m} \in [0, 1]$ denotes predicted affordance map values and ${m} \in [0, 1]$ denotes the ground-truth propagation score (0 is the minimum value of correlation for each affordance label, 1 is the maximum value of correlation). $L_{ce}$ is the cross-entropy loss and dice loss $L_{dice}$ is also introduced to mitigate the imbalance issue caused by the dataset. In our training process, hyper parameters $W_1$ and $W_2$ are set as 0.4 and 0.6, separately.
\section{Experiments}
\subsection{Implementation details}
\textbf{Data Augmentation:} The object point cloud were captured through rotating at a random Euler angle $(x, y, 0)$, where the range of rotation is $x \in [0, 2\pi], y \in [-\pi/2, \pi/2]$. The rotation number of each object is 900 times, where jitter and dropout operation are added in each rotation. After that, the observed partial point cloud is sampled to 2048 points through the farthest point sampling (FPS) algorithm during training. And they are also processed through a mean centered and unit-scaled trick. For the visual affordance prediction, where we followed \cite{deng20213d}, the point cloud needs to be further normalized during the training.

\textbf{Fine Grasp Candidate Generation:} For each affordance label $i$, all coarse grasp candidates $\mathcal{G_C}^i$ output by the grasping detection module are sorted in descending order according to theirs confidence scores $\mathcal{S_C}^i$. Then, the predicted values of affordance heatmap from visual affordance module can be also obtained and sorted descending as $H^N_i$. To reduce the computational cost for the sampling sparse point cloud, we take the point cloud area where the top 100 maximum values in the $H^N_i$ are located (outliers are filtering) to approximately represent our affordance labels, which could be expressed as $(\mathcal{P}^{100}_i, H^{100}_i)$. Note that most of values of $H^N_i$ are approximate 0 because they do not belong to label $i$. After that, each point from $\mathcal{P}^{100}_i$ is selected to compute the L2 distance with the middle control point  $P_i^{cm}$ of each coarse candidate, the minimal distance is regarded as the vision-guided score for each grasp:
\begin{equation}
    \mathcal{S_V}^i = min(||\mathcal{P}^{j}_i - P_i^{cm}||_2), \text{for } j \in [0,100)
\end{equation}

The final grasp candidate can be obtained by combining the value of $\mathcal{S_C}^i$ and $\mathcal{S_V}^i$: 
\begin{equation}
    \mathcal{S_F}^i =\alpha_1 * \mathcal{S_C}^i + \alpha_2 *\mathcal{S_V}^i
\end{equation}
where $\mathcal{S_F}^i$ represents the score of fine grasp candidates for affordance label $i$ and  the hyperparameters of $\alpha_1$, $\alpha_2$ are both set to 0.5.

\subsection{Quantitative Evaluation of Proposed Framework}
We design two baselines for comparison with our method. 1) Baseline1: This is from the similar work\cite{murali2020same}, which takes the scanned point cloud as input to train a framework to generate different task-oriented grasps.  The grasp detection benchmark from this work is adapted from \cite{mousavian20196} and knowledge graph is introduced to  connect diverse tasks and objects. However, we find that the effect of knowledge graph is limited in our dataset due to fewer tasks and object categories. 2) Baseline2: This can be considered as a degraded version of our method, where the multi-stream implicit estimation method is employed to detect grasp without the guiding of the 3D visual affordance.  To demonstrate the effectiveness of our method, we also design an evaluation metric to compare the grasp similarity between predicted grasp and ground truths in our dataset. Similar to the loss function of the implicit estimation network, we sample 100 predicted grasps and ground truth grasps randomly and then we will compare the L1 distance of each predicted grasp $\mathcal{G}$ with $\mathcal{G(T)^*}$, the minimum value is assumed as its similarity value. The mean value of the sum of 100 minimum values can be computed as the evaluated similarity metric (ESM). And a smaller ESM value means a better similarity between predicted grasps and ground truths.

\subsubsection{Ablation Study of grasping affordance detection} 
\begin{figure}[t!]
\centering
\includegraphics[width=0.95\linewidth]{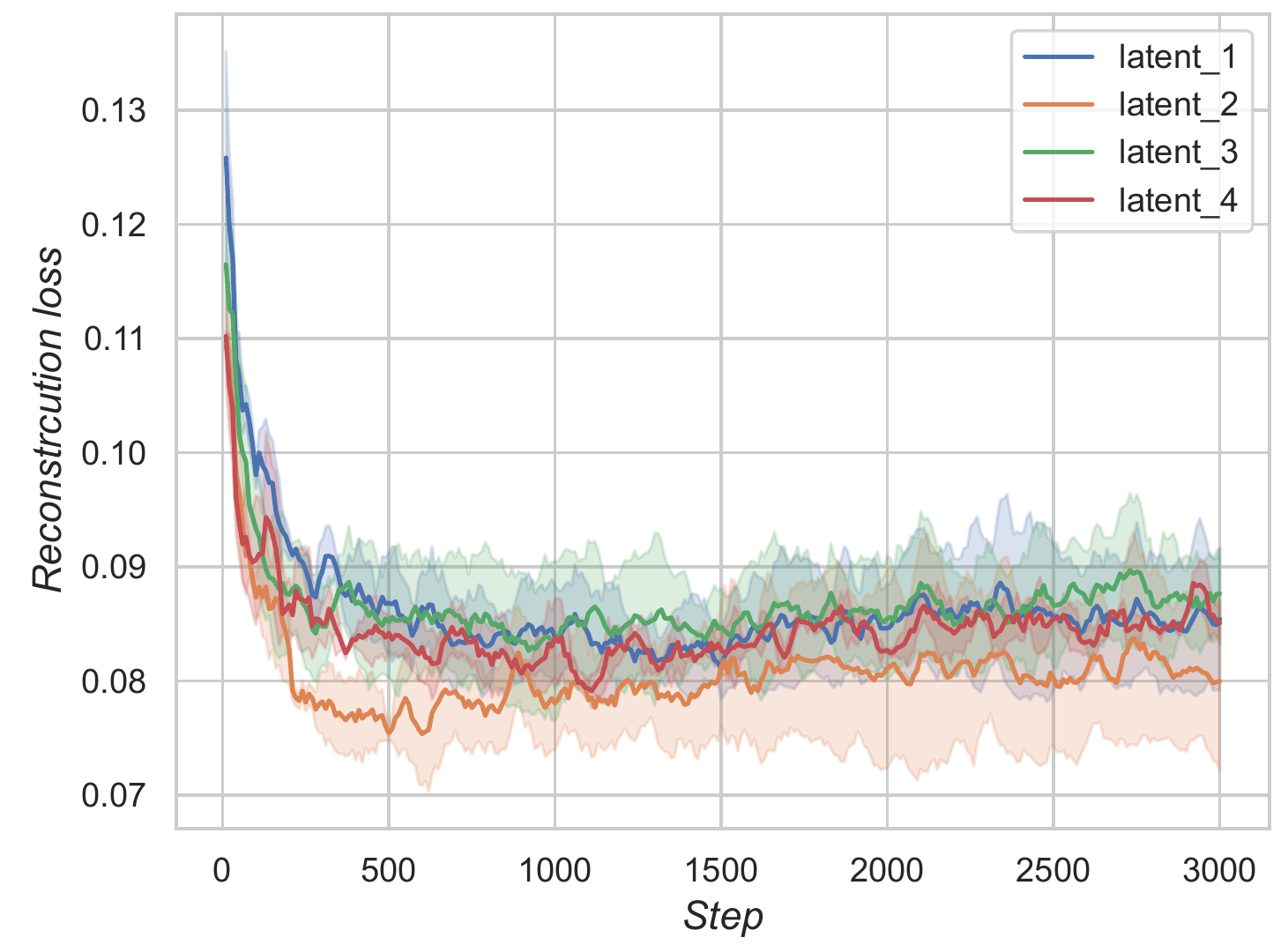}
\caption{The effect of length of latent vector in our implicit estimation network.
}
\label{latent_effect}
\vspace{-0.01cm}
\end{figure}

We study the effects of the grasp detection model from our implicit estimation network (IEN) and widely used VAEs model\cite{mousavian20196}, and the comparison results of ESM are listed in Table~\ref{tab:implicit_model}. It shows that our network can achieve a better prediction result than the VAEs model. Moreover, we illustrate the effect of length of the latent vector in our implicit estimation network. As can be seen from Figure~\ref{latent_effect}, when the length of latent vector equals to 2, the network achieves the best performance in the test set because a slightly bigger latent vector can cause the over-fitting problem.  

\begin{table}[t!]
\caption{Comparison of different generators: IEN and VAEs}
\resizebox{\linewidth}{!}{%
\begin{tabular}{l|c|c|c|c|c|c|l}
\hline
\textbf{ESM} & \textbf{Mug} & \textbf{Bottle} & \textbf{Bowl} & \textbf{Hat} & \textbf{Scissors} & \textbf{Knife} & \textbf{Average} \\ \hline
IEN                 & 0.062        & 0.058           & 0.081         & 0.122        & 0.157             & 0.071          & 0.092            \\ \hline
VAEs                  & 0.106        & 0.112           & 0.142         & 0.186        & 0.211             & 0.147          & 0.151            \\ \hline
\end{tabular}}
\label{tab:implicit_model}
\end{table}

\subsubsection{Ablation Study of the visual affordance module}
As the vision-guiding module, the visual affordance network is also the most important component of our framework. Thus, we continue to study the effects of various designs existing in visual affordance network. Though point-level (PointNet\cite{qi2017pointnet} and PointNet++\cite{qi2017pointnet++}) and edge-level (DGCCN\cite{wang2019dynamic}) method are used in the benchmark~\cite{deng20213d}, we denote the levels of new features, i.e. point feature, edge feature, and MHSA module, respectively. The results of three variants are listed in Table~\ref{tab:visual_ablation}.  Eventually, the integration features of three levels give us the best performance on the visual affordance prediction.

\begin{table}[b]
\caption{Ablation study for the visual affordance module}
\resizebox{\linewidth}{!}{%
\label{tab:visual_ablation}
\begin{tabular}{|c|c|c|c|c|c|c|}
\hline
Point feature & Edge feature & MHSA module & average AP  & average AUC & average IOU\\
\hline

\XSolidBrush & \CheckmarkBold & \CheckmarkBold & 0.4201   & 0.8325   & 0.1098 \\
\CheckmarkBold & \XSolidBrush & \CheckmarkBold & 0.3737   & 0.7982   & 0.1473 \\
\CheckmarkBold & \CheckmarkBold & \XSolidBrush & 0.4241   & 0.8249   & 0.1618 \\
\CheckmarkBold & \CheckmarkBold & \CheckmarkBold & \textbf{0.4281}   & \textbf{0.8360}   & \textbf{0.1628} \\
\hline
\end{tabular}}

\end{table}

\subsubsection{Comparison with baselines}
Table~\ref{tab:existing} and ~\ref{tab:novel} summarize the ESM results of comparing our method to the baselines on existing objects and novel objects, respectively. It is not surprising that the using of implicit representation leads to improvements for task-oriented grasp prediction. Moreover, for most object point clouds, the 3D affordance map in our method could effectively improves the final prediction result. We also observe a phenomenon that the coarse version of our approach (Baseline2) sometimes is better than our method. This is probably because the input partial point cloud sometimes lacks the affordance context. For example, \emph{grasp} of \emph{mug} corresponds to the mug handle. If the captured mug point cloud misses the handle completely, that will cause a bad affordance heatmap. Our method also shows that the best performance for novel objects, demonstrating a great ability for generalization.

\begin{table}[t]
\caption{Evaluated similarity metric from different existing objects for different-oriented tasks}
\resizebox{\linewidth}{!}{%

\begin{tabular}{c|cccccc|cccccc}
\hline
\multirow{2}{*}{\textbf{Task}}            & \multicolumn{6}{c|}{\textbf{Grasp}}                                                                                                                                   & \multicolumn{6}{c}{\textbf{Wrap/Cut}}                                                                                                                               \\ \cline{2-13} 
                                 & \multicolumn{1}{c|}{Mug1}  & \multicolumn{1}{c|}{Mug2}  & \multicolumn{1}{c|}{Mug3}  & \multicolumn{1}{c|}{Bottle1} & \multicolumn{1}{c|}{Knife1} & Scissor1 & \multicolumn{1}{c|}{Mug1}  & \multicolumn{1}{c|}{Mug2}  & \multicolumn{1}{c|}{Mug3}  & \multicolumn{1}{c|}{Bow1}  & \multicolumn{1}{c|}{Knife1} & Scissor1 \\ \hline
Baseline1           & \multicolumn{1}{c|}{0.167} & \multicolumn{1}{c|}{0.147} & \multicolumn{1}{c|}{0.107} & \multicolumn{1}{c|}{0.149}   & \multicolumn{1}{c|}{0.151}  & 0.321    & \multicolumn{1}{c|}{0.141} & \multicolumn{1}{c|}{0.088} & \multicolumn{1}{c|}{0.096} & \multicolumn{1}{c|}{0.203} & \multicolumn{1}{c|}{0.185}  & 0.185    \\ \hline
Baseline2          & \multicolumn{1}{c|}{0.061} & \multicolumn{1}{c|}{0.051} & \multicolumn{1}{c|}{0.048} & \multicolumn{1}{c|}{0.092}   & \multicolumn{1}{c|}{0.138}  & 0.248    & \multicolumn{1}{c|}{0.097} & \multicolumn{1}{c|}{0.094} & \multicolumn{1}{c|}{0.084} & \multicolumn{1}{c|}{0.133} & \multicolumn{1}{c|}{0.133}  & 0.139    \\ \hline
Our & \multicolumn{1}{c|}{0.046} & \multicolumn{1}{c|}{0.032} & \multicolumn{1}{c|}{0.033} & \multicolumn{1}{c|}{0.110}   & \multicolumn{1}{c|}{0.083}  & 0.179    & \multicolumn{3}{c|}{---}                                                             & \multicolumn{1}{c|}{0.121} & \multicolumn{1}{c|}{0.117}  & 0.144    \\ \hline
\end{tabular}}
\label{tab:existing}
\end{table}

\begin{table}[t]
\caption{Evaluated similarity metric from different novel objects for different-oriented tasks}

\resizebox{\linewidth}{!}{%

\begin{tabular}{c|ccclcc|ccclcc}
\hline
\multirow{2}{*}{\textbf{Task}}            & \multicolumn{6}{c|}{\textbf{Grasp}}                                                                                                      & \multicolumn{6}{c}{\textbf{Wrap/Cut}}                                                                                                   \\ \cline{2-13} 
                                 & \multicolumn{1}{c|}{Mug4}  & \multicolumn{1}{c|}{Mug5}   & \multicolumn{2}{c|}{Scissor2} & \multicolumn{1}{c|}{Knife2} & Knife3 & \multicolumn{1}{c|}{Mug4}  & \multicolumn{1}{c|}{Mug5}  & \multicolumn{2}{c|}{Scissor2} & \multicolumn{1}{c|}{Knife2} & Knife3 \\ \hline
Baseline1           & \multicolumn{1}{c|}{0.130} & \multicolumn{1}{c|}{0.136}  & \multicolumn{2}{c|}{0.192}    & \multicolumn{1}{c|}{0.156}  & 0.178  & \multicolumn{1}{c|}{0.082} & \multicolumn{1}{c|}{0.109} & \multicolumn{2}{c|}{0.234}    & \multicolumn{1}{c|}{0.185}  & 0.195  \\ \hline
Baseline2          & \multicolumn{1}{c|}{0.117} & \multicolumn{1}{c|}{0.056} & \multicolumn{2}{c|}{0.161}    & \multicolumn{1}{c|}{0.091}  & 0.122  & \multicolumn{1}{c|}{0.076} & \multicolumn{1}{c|}{0.101} & \multicolumn{2}{c|}{0.187}    & \multicolumn{1}{c|}{0.131}  & 0.162  \\ \hline
Our & \multicolumn{1}{c|}{0.059} & \multicolumn{1}{c|}{0.037}  & \multicolumn{2}{c|}{0.161}    & \multicolumn{1}{c|}{0.052}  & 0.104  & \multicolumn{2}{c|}{---}                                & \multicolumn{2}{c|}{0.162}    & \multicolumn{1}{c|}{0.104}  & 0.134  \\ \hline
\end{tabular}}
\label{tab:novel}
\end{table}

\begin{figure}[t!]
\centering
\includegraphics[width=1.0\linewidth]{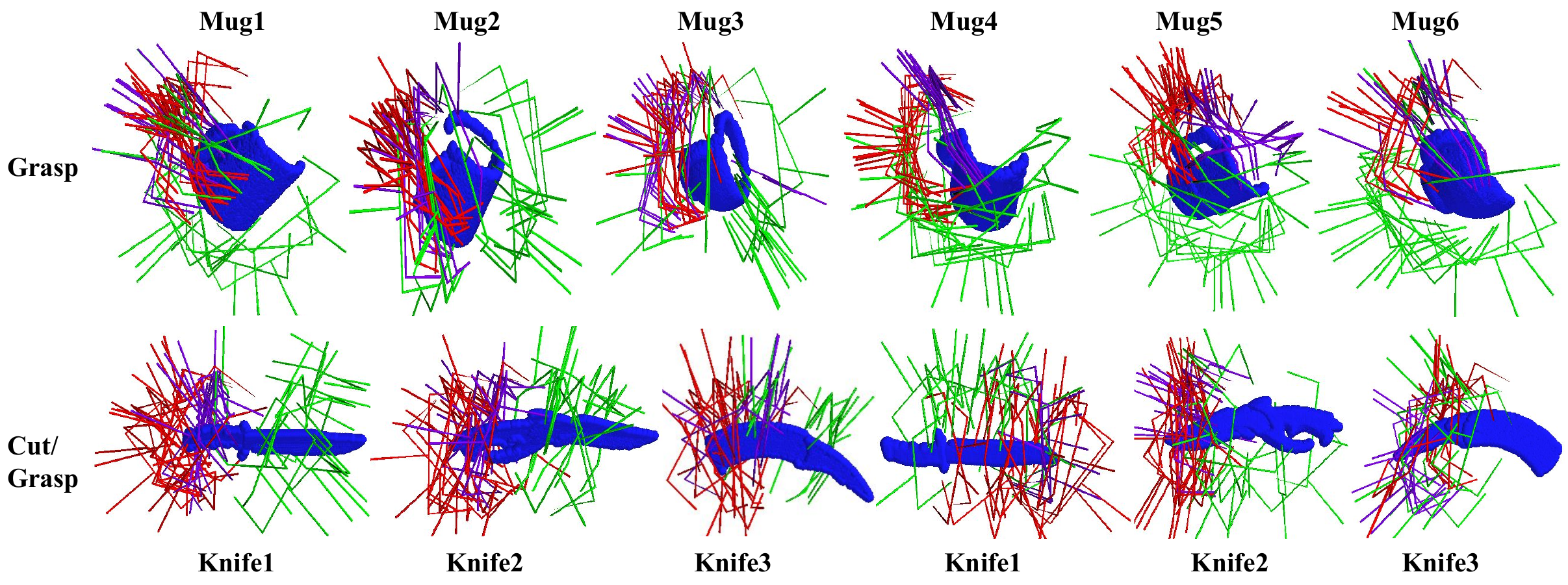}
\caption{Visualization results of our proposed method for task-oriented grasp prediction. For each affordance label, green means the generated grasp without visual guidance, purple means grasp candidates from our method, and red means the ground truth from our dataset.
}
\label{visualization_grasp}
\vspace{-0.1cm}
\end{figure}

\subsubsection{Visualization Analysis}
\begin{figure}[t!]
    \centering
    \includegraphics[width=0.3\textwidth]{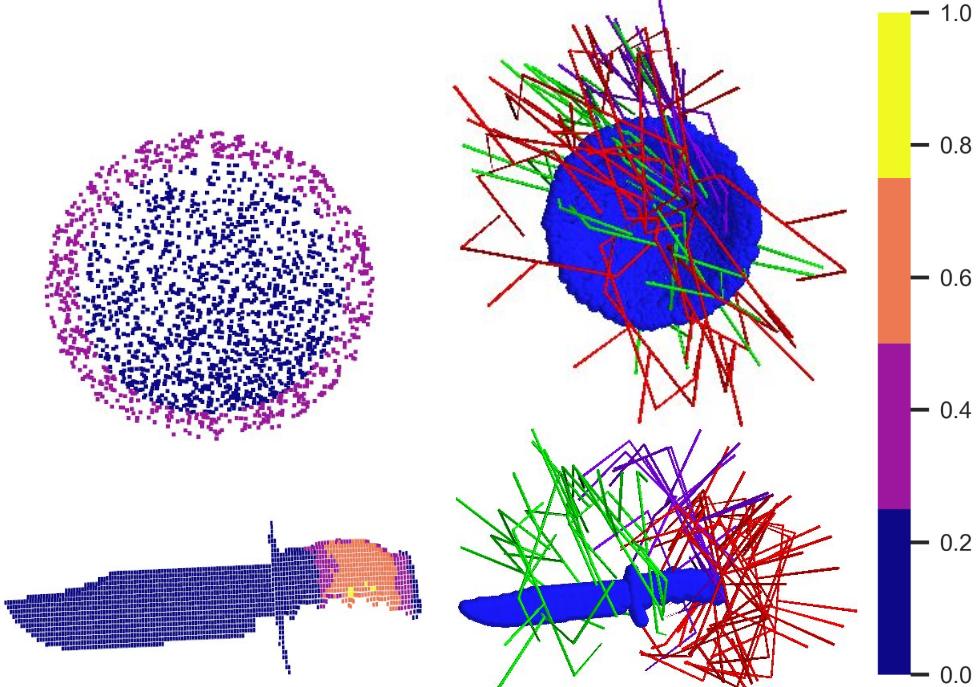}
    \caption{Visualization results of the 3D affordance map and corresponding grasp candidates (\emph{wrap} of the bowl and \emph{cut} of the knife).
    }
    \label{simulation}
    \vspace{-0.1cm}
\end{figure}

Figure~\ref{visualization_grasp} and~\ref{simulation} show the visualization results of our proposed method for task-oriented grasp prediction, respectively.  The results of affordance-based grasp candidates from our method  are compared with the predictions from the degraded version and ground truths. Seen from Figure~\ref{visualization_grasp}, it is very challenging to predict the grasps restricted in a small affordance region. The coarse grasp candidates from degraded version of our method are not ideal because many predicted grasps are not 
centric-around the affordance context (like \emph{grasp} around the handle of a mug, and \emph{cut} around the handle of a knife). However, the predictions from our method could be more concentrated on the affordance area, and the fine grasp candidates are more accurate. Figure~\ref{simulation} also shows the visualization results of the 3D affordance map and corresponding
grasp candidates. We find an interesting result when the affordance region is evenly distributed over the contour of the object (like \emph{wrap} of the bowl and mug), where the predictions of our method will be similar to the coarse candidates. That is because the groundtruths are uniformly distributed around the whole point cloud, causing the vision-guided score $\mathcal{S_V}^i$ cannot effectively improve the original grasping score $\mathcal{S_C}^i$.

\subsection{Real Robotic Evaluation}
\begin{figure}[t!]
    \centering
    \includegraphics[width=0.3\textwidth]{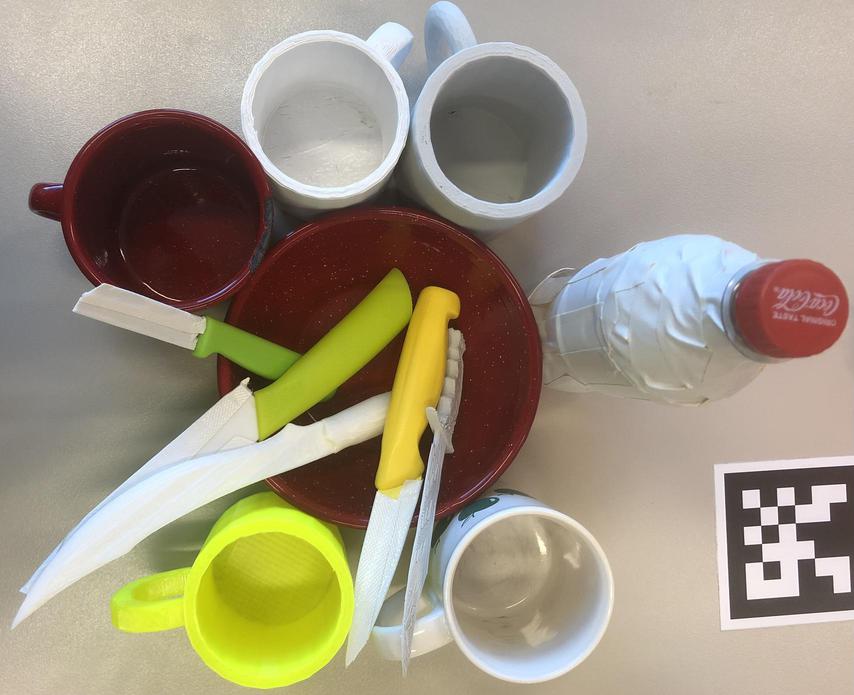}
    \caption{All novel objects that are tested in our real robot experiment.
    }
    \label{objects}
    \vspace{-0.1cm}
\end{figure}

\begin{figure}[t!]
    \centering
    \includegraphics[width=\linewidth]{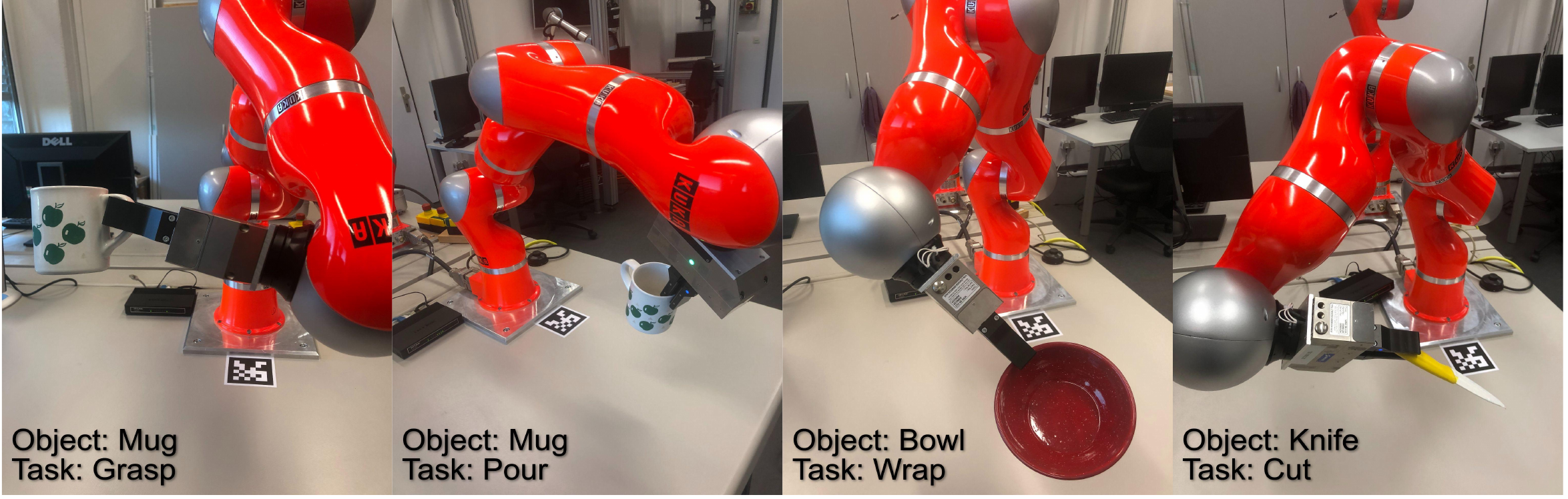}
    \caption{Evaluation results based on different tasks and objects from real robot grasping.
    }
    \label{experiments}
    \vspace{-0.1cm}
\end{figure}

To evaluate the performance using our task-oriented grasping detection framework for robotic grasping, we run real robot experiments to demonstrate that our model trained from simulation transfers well to the real robot environment. In the experiment setup, we put a single object on a flat table at an arbitrary pose without any clutter. As shown in Figure~\ref{objects}, all the objects we test are unknown to the system. To avoid causing damage to real objects and gripper, some objects like mug and knife are obtained through 3D-print technology. Near the target object, a KUKA LWR robot is fixed on the table with a 2-fingered WSG-50 gripper. And about 1.2\,m in front of the robot and target object, a Mechmind RGB-D camera is suspended on a bracket to capture the object point cloud. The obtained point cloud is input into the trained framework. For each task, over 3000 successful coarse grasp candidates are output first. Then a 3D visual affordance map for this affordance label is predicted to couple with the original grasp score. Finally, 20 fine-sampled grasp candidates are sent to the robot for execution according to their final evaluated grasping scores in descending order, where the first grasp (highest score) will be executed if there is no problem for its planning in MoveIt. The whole framework is trained and inferenced on the desktop PC with NVIDIA GTX 2080Ti GPU. Figure~\ref{experiments} shows the evaluation examples based on different tasks and objects from real robot grasping. It demonstrates that our approach trained in simulation can be validated successfully in a real environment. During the process of experiments, we find that if more affordance features like handle of mug and knife can be captured by camera, the final execution grasp is more stable and accurate. And the gap between simulation and real environment also exists when the texture of real object is very smooth and irregular though the generate grasp looks good. Nonetheless, for almost each experiment, our fine grasp poses mainly focus on corresponding affordance area, showing our framework can well reason the relationship between grasp detection and task-oriented affordance.

\section{Conclusion and Future Work}
This paper investigates the challenging problem of task-oriented robotic grasping. Focusing on 6-DoF grasp detection, we proposed a novel solution through designing three modules: an implicit estimation network, a grasp evaluation network, and an attention-aware visual affordance network, which achieves consistent and clear improvements over baselines for existing and novel objects in our self-constructed affordance grasp dataset. This work provides several key insights into task-oriented grasping: 1) the learning of implicit representations from object and grasp poses from each affordance label is the core of 6-DoF grasping affordance detection. 2) the exploitation of point-based, edge-based features and the attention mechanism are necessary to achieve a better affordance map prediction. 3) the generated 3D affordance map could effectively guides coarse grasp candidates to become more accurate and finer for a specific affordance task. For the future work, we hope to use the advantage of the simulation environment to rapidly extend the number of object categories and affordance labels, improving the generalization ability of our framework. Moreover,  we also want to explore the possibility to combine the trained framework with hand-over task, which is beneficial to increase the use of affordance learning in robotic field.

\section{Acknowledgements}
This research was funded by the German Research Foundation (DFG)
and the National Science Foundation of China (NSFC)
in the project Crossmodal Learning, DFG TRR-169/NSFC 61621136008, in project
DEXMAN under grant No.410916101, and partially supported by European projects H2020 Ultracept (778602).
We would like to thank Norman Hendrich, Chao Zeng and Yuting Sun for their generous help and insightful advice.
We also thank Mech-Mind Robotics Company for providing the 3D camera.

\bibliographystyle{IEEEtran}
\bibliography{bibtex/bib/reference}

\end{document}